\documentclass[12pt]{article}%
\usepackage{amsfonts}
\usepackage{geometry}
\usepackage[final]{graphicx}
\usepackage{verbatim}
\usepackage[]{graphics}
\usepackage{epsfig}
\usepackage{amsmath}
\usepackage{amssymb}
\usepackage{amsthm}
\usepackage{lscape}
\usepackage{setspace}
\usepackage{etoolbox}
\apptocmd{\sloppy}{\hbadness 10000\relax}{}{}
\usepackage[hyphens]{url}
\usepackage[bottom]{footmisc}
\usepackage{endnotes}
\usepackage{setspace}
\usepackage{float}
\usepackage{xcolor}%
\providecommand{\U}[1]{\protect \rule{.1in}{.1in}}
\begin{document}

\author{Marc Maliar}
\title{How Machine (Deep) Learning Helps Us Understand Human Learning: the Value of
Big Ideas\thanks{University of Chicago. I started this project while interning
at AgilOne tech startup \protect\url{https://www.agilone.com} -- I am grateful to my mentors and colleagues for their help and encouragement, Vlad Kozlov, Rajiv Shringi,
Jacek Sniecikowski, Ajit Sutar, among others. I thank Prof. Han Hong (Stanford
University) and Prof. David Laibson (Harvard University) for useful comments
and suggestions. All errors are mine. }}
\date{February 10, 2019}
\maketitle

\begin{abstract}
I use simulation of two multilayer neural networks to gain intuition into the
determinants of human learning. The first network, \textit{the teacher}, is
trained to achieve a high accuracy in handwritten digit recognition. The
second network, \textit{the student}, learns to reproduce the output of the
first network. I show that learning from the teacher is more effective than
learning from the data under the appropriate degree of regularization.
Regularization allows the teacher to distinguish the trends and to deliver
"big ideas" to the student. I also model other learning situations such as
expert and novice teachers, high- and low-ability students and biased learning
experience due to, e.g., poverty and trauma. The results from computer
simulation accord remarkably well with finding of the modern psychological
literature. The code is written in MATLAB and will be publicly available from
the author's web page.

\qquad

\qquad

\qquad \newline$Key$ $Words:$ deep learning, neural networks, machine learning,
human learning\newpage

\end{abstract}

\section{Introduction}

Nowadays, it is hard to find an area of human life where artificial
intelligence would not have its applications, starting from composing music
and writing novels to self-driving cars and anti-spam filters. As is evident
from the term \textit{machine learning}, the main goal of the research in
artificial intelligence is to teach machines. "Artificial neurons" were
designed to perform logical operations in the same way humans do. But the
learning process can go in the opposite direction as well. In particular,
brain scientists believe that we can get a better understanding of how humans
learn by studying how machines learn.\footnote{In spite of similarities,
researchers found some important differences between humans and machines.
Specifically, our bodies help our memories; metaphors are powerful learning
tools; there is no substitute for learning-by-doing; good teaching draws on
shared experience; see Briggs (2017) for a discussion.} Researchers, educators
and students can benefit from using machine learning to explain the learning
process undertaken by humans.

Human learning can be studied in many different ways (e.g., by analyzing the
existing empirical data, by running lab experiments, by looking at the
highlighted areas of a human brain). Recent developments in deep learning have
opened new possibilities for studying human learning by means of computer
simulation; see Domingos (2015). This is the approach I follow in the present paper.

I use deep-learning neural networks to analyze how the design of human
learning process can influence its outcome. One important question is under
what conditions do humans learn better from teachers than from their own
experience (represented by raw data from the real world). I address these
questions in the context of an image recognition problem, namely,
classification of handwritten numbers. Handwritten character recognition is
one of the most challenging problems in the field of pattern recognition; see
Purohit and Chauhan (2016) for a survey of the literature on handwritten
character recognition. The techniques used in this field rely on standard deep
learning; see Goodfellow et al. (2016) and Hastie et al. (2008). Industry
developments in software, e.g., TensorFlow software, make it feasible to
handle problems of very large dimensionality; see Hope et al. (2017).

The main novelty of the present paper is the way of modeling the interaction
between the teacher and student, which I propose in the context of
classification problem. Learning from a teacher means that a previously
trained neural network (teacher) observes an image and produces probabilities
that the given image corresponds to digits 0, 1, ..., 9. These probabilitities
are provided as input to train the student network. On opposite, a student who
learns from its own experience observes the handwritten digit images and has
no other information. I design the learning environment to mimic several real
world learning scenarios in which I vary the number of training samples
(expert versus novice teacher), regularization level (the teacher's ability to
categorize information into big ideas) and the selection of the training
samples (constructing "bad" teacher with a biased perception of the learning
process). I also vary the student's learning rate (representing the ability of
students) and the selection of training samples (modeling the students with
atypical learning experience due to, e.g., poverty and trauma). To measure the
success of learning, I report two statistics: the predictive accuracy of image
recognition and the cost on a test set. The code is written in MATLAB and will
be publicly available on my web site.

My findings are as follows: (i) If the teacher is sufficiently trained, then a
student with a high learning rate learns faster from the teacher than from the
data and attains a higher accuracy of digit recognition; in contrast, a
student with a low learning rate does not significantly benefit from the
teacher. (ii) The success of learning depends on the qualification of the
teacher relative to the qualification of the student; overqualification of the
teacher does not improve the learning outcomes but underqualification worsens
such outcomes; (iii) Learning from a "bad" teacher \textbf{(}i.e., with a
biased perception of the learning process) is less effective than learning
from the data, and regularization only increases the bias. (iv) Low-ability
students benefit less from a teacher than high-ability students. (v) Learning
from an unbiased teacher can correct biased views of the students derived from
their atypical experience (e.g., poverty and trauma); and moderate
regularization increases the effectiveness of the teacher.

The main novelty and methodological contribution of the present paper is to
show how to model the learning of one artificial intelligence from another. My
analysis highlights the importance of "big ideas" for successful learning: a
student network progresses most in my experiments when the teacher's input is
sufficiently regularized. The degree of regularization determines how well the
teacher is able to distinguish and generalize the regularities in the data.
This finding accords with the conclusion of modern psychological literature
about the importance of learning experiences which specifically enhance the
students' abilities to recognize meaningful patterns of information; see
Section 2 for a survey of the related literature. The knowledge of experts is
not simply a list of facts and formulas that are relevant to their domain;
instead, their knowledge is organized around core concepts or
\textquotedblleft big ideas\textquotedblright \ that guide their thinking about
their domains.

The paper is organized as follows: Section 2 describes the evidence on the key
determinants of successful learning, Section 3 outlines the methodology of my
computer simulation. Section 4 investigates how the teacher's characteristics
affect the learning outcomes. Section 5 discusses how the learning outcome
depends on the student's characteristics. Finally, Section 7 concludes.

\section{Evidence on human learning}

Learning plays a critical role in both personal development of each individual
and the evolution of the society as a whole. Some relevant evidence on human
learning is summarized below.

\paragraph{i) The critical role of the teacher in a student's learning
outcomes.}

There is a large body of modern psychological literature that emphasizes the
critical role of a teacher in successful learning outcomes. For example,
Hattie (2003) found that excellence in teaching is the single most powerful
influence on achievement and found that teachers account for 30\% of the
variance in the students' learning outcomes. Ulug et al. (2011) concluded that
teachers' positive attitudes have positive effects on students' performance
and personality developments and are the second-highest determining factor in
the development of individuals, after parents. Finland is often cited to have
the best educational system in the world and the key factor for its success
seems to be the training of teachers: Finland's teachers are required to have
Master's degrees, and teaching careers are the most competitive in the country.

\paragraph{ii) Expert teachers and the importance of big ideas.}

But what does it mean to be an expert teacher? Bransford, Brown, and Cocking,
(2000) survey the literature investigating how teachers-experts differ from
teachers-novices in such areas as chess, physics, mathematics, electronics,
and history. Their conclusion is that experts' knowledge is organized around
important "big" ideas that lead to deeper conceptual understanding of their
domain. As an illustration, consider an experiment from DeGroot (1965). A
chess master and a novice player were given 5 seconds to memorize a chess
board position. The master was able to reconstruct the position far more
accurately than the novice but only when the chess pieces were arranged in
configurations that conformed to meaningful games of chess; when the pieces
were randomized both the master and novice had similar recall. Hatano and
Inagaki (1986) argue that the key to effective learning is adaptive expertise:
The experts not only use their current knowledge but attempt to continually
improve their expertise. The main challenge for theory of learning is
therefore to understand how the "virtuosos\textquotedblright \ organize their
knowledge into big ideas.

\paragraph{iii) Bad teachers are insufficiently trained and have a biased
perception of the learning process.}

Another piece of evidence is provided by Foote et al. (2000) who conducted a
questionnaire about secondary school teachers. One common response amongst
students, teachers, administrators and parents was that bad secondary teachers
do not posses the knowledge of their discipline in the degree that is
sufficient for effective teaching. The other common response was that bad
teachers do not have adequate perception of the learning process, making their
lessons too fast, slow, easy, or difficult. In other words, the teacher's
perception (experience) is biased in the sense that it does not accord well
with the learning needs of the students.

\paragraph{iv) High-ability students benefit from enriched teaching but not
low-ability students.}

There is also evidence on learning outcomes by students groups. Kulik and
Kulik (1982) summarize the results from 52 studies of grouping students by
their abilities in secondary schools. One robust finding was that studies in
which high-ability students received enriched instructions in honors classes
produced significant learning improvements, while studies of average and
below-average students produced near-zero effects. In other words, only high
ability students benefited from more advanced teaching, in contrast to low
ability students.

\paragraph{iv) Students can learn from the educator what they missed in their
own experience.}

Finally, there is abundant evidence about learning of students from low
socioeconomic backgrounds. The handbook by Izard (2016) edited by the US
National Educational Association states: "Because students impacted by poverty
and trauma have not learned appropriate emotional responses, when the educator
models appropriate social behaviors, the student's mirror neurons that were
neglected or harmed earlier can still pick up on clues from the educator to
learn now what they missed earlier". That is, the educator can become the
missing person and fill in the socioemotional gap for those students whose own
perception is affected by the negativity of poverty and trauma.

\section{Modeling the learning process}

In this section, I present the methodology of our analysis. I first describe
the topology of the neural network studied, then explain how the teacher and
students are trained, and finally, show how our computer simulation is
connected to the evidence on human learning.

\subsection{Topology of neural network}

All neural networks studied in this paper have the same topology. They have
three layers, $n$ =\ 1, 2, and 3: the input layer ($400\times25$), one hidden
layer ($25\times10$), and the output layer ($10\times1$). Thus, the number of
nodes in the three layers are $400$, $25$ and $10$, respectively. The output
layer returns a $\left(  10\times1\right)  $ vector; ten elements in this
vector represent the probabilities that the image is a digit $0,1,...,9$, respectively.

The neural network's activation function is a sigmoid function given by%
\[
\sigma \left(  x\right)  =\frac{1}{1+e^{-x}},
\]
for any $x\in%
\mathbb{R}
$. The cost function is given by%
\begin{equation}
J\left(  \theta \right)  =\frac{1}{m}%
{\displaystyle \sum \limits_{i=1}^{m}}
{\displaystyle \sum \limits_{k=1}^{10}}
\left[  -y_{k}^{i}\log(h_{k}\left(  \theta,X^{i}\right)  )-(1-y_{k}^{i}%
)\log(1-h_{k}\left(  \theta,X^{i}\right)  )\right]  +\frac{\lambda}{2m}%
{\displaystyle \sum \limits_{j=1}^{10,285}}
\theta_{j}^{2}, \label{C}%
\end{equation}
where $m$ is the number of training samples, $X^{i}\in%
\mathbb{R}
^{400\times1}$ is the input image and $y^{i}\in%
\mathbb{R}
^{10\times1}$ is the actual output vector for a sample $i$, $\theta \in%
\mathbb{R}
^{10,285\times1}$ is the vector of weights of the neural network, $h:%
\mathbb{R}
^{400\times10,285}\rightarrow%
\mathbb{R}
^{10\times1}$ is the predicted output vector, and $\lambda \in \left[
0,\infty \right)  $ is the regularization parameter; see Goodfellow et al.
(2016) for details.

We use back propagation\ to calculate the gradient of every node. These values
are used to train the networks. The partial derivative of the $n$th layer for
weight matrix $\theta_{n\text{ }}$is%
\[
\frac{\partial J}{\partial \theta_{n}}=%
{\displaystyle \sum \limits_{i=1}^{m}}
\delta_{n+1}^{i}a_{n}^{i}+\frac{\lambda}{m}\theta_{n},
\]
where $a_{n}^{i}$ is the activated output, and $\delta_{n}^{i}$ is the error
for the current layer $n$ and sample $i$. For the last layer $n$ = 3,
$\delta_{n}^{i}\in%
\mathbb{R}
^{10\times1}$ is%
\[
\delta_{3}^{i}=h\left(  \theta,X^{i}\right)  -y^{i},
\]
where $h\left(  \theta,X^{i}\right)  $ is the output of the neural network,
and $y^{i}$ is the real digit vector, both for sample $i$.

For all other layers, i.e., $n=1,2$, $\delta_{n}^{i}$ is%
\[
\delta_{n}^{i}=\left(  \delta_{n+1}^{i}\times \theta_{n}\right)  .\times \left[
\sigma(z_{n}^{i}).\times(\sigma(z_{n}^{i})-1)\right]  ,
\]
where $\theta_{n}$ is the matrix containing the weights for layer $n$,
$\sigma$ is the sigmoid function, and $z_{n}^{i}$ is the inactivated output of
layer $n$ ($z_{n}^{i}=\theta_{n}\times a_{n-1}^{i}$) for sample $i$. The
operator $.\times$ represents an element-wise matrix multiplication, and the
operator $\times$ is a matrix multiplication.

\subsection{The learning process}

In this section, I describe the data, as well as the training procedures for
the teacher and students.

\subsubsection{Data}

We\ train the neural networks with a handwritten digit classification problem
$20\times20$ pixel images---these images come from the database provided by Ng
(2017). We have 5,000 total training samples composed of $\left(  X^{i}%
,y^{i}\right)  $, where $X^{i}\in%
\mathbb{R}
^{400\times1}$, $y^{i}\in%
\mathbb{R}
^{10\times1}$, for sample $i$ from $1$ to $m$. We begin by randomizing the
order of the samples. Then, we set apart 1,000 samples as the test set. These
samples will be used to evaluate the performance of neural networks later. No
neural network will have access to these samples. Then, we take 500 samples
for training the students, and we call this the student training set
$(X_{student},y_{student})$. To train the teacher, we use (up to) 3,500 of the
remaining samples $(X_{teacher},y_{teacher})$. The teacher training set and
the student training set will never overlap.

\subsubsection{The criteria of learning success}

As a criteria of the learning success, we consider two measurements:\ accuracy
(number of samples predicted correctly on a test set) and cost (the value of
cost function (\ref{C}) on the test set, which is a real number showing how
far off the predictions are from the correct answers). To reduce sampling
errors, I report the average accuracy and cost which are averaged over ten simulations.

\subsubsection{Training the teacher network}

We train the teacher on the teacher's training set $(X_{teacher},y_{teacher})$
by minimizing the cost function (\ref{C}). We use the MATLAB\ optimization
routine \textit{fmincg} that guarantees convergence: we don't want to study
how quickly the teacher learns but we want the teacher to learn as much as
possible from the data available (in contrast, for students, we will use
stochastic gradient which allows us to investigate the learning speed). In
Figure 1, I show the accuracy and cost of the trained teacher for 4 training
sets consisting of 500, 1000, 2000 and 4000 handwritten samples (I connect the
3 points on the graph with a line for expositional convenience); we vary the
learning rate $\lambda \in \left \{  0,5,10\right \}  $.%

\begin{figure}[H]

\begin{center}
\includegraphics[
height=3.17in,
width=6.0287in
]%
{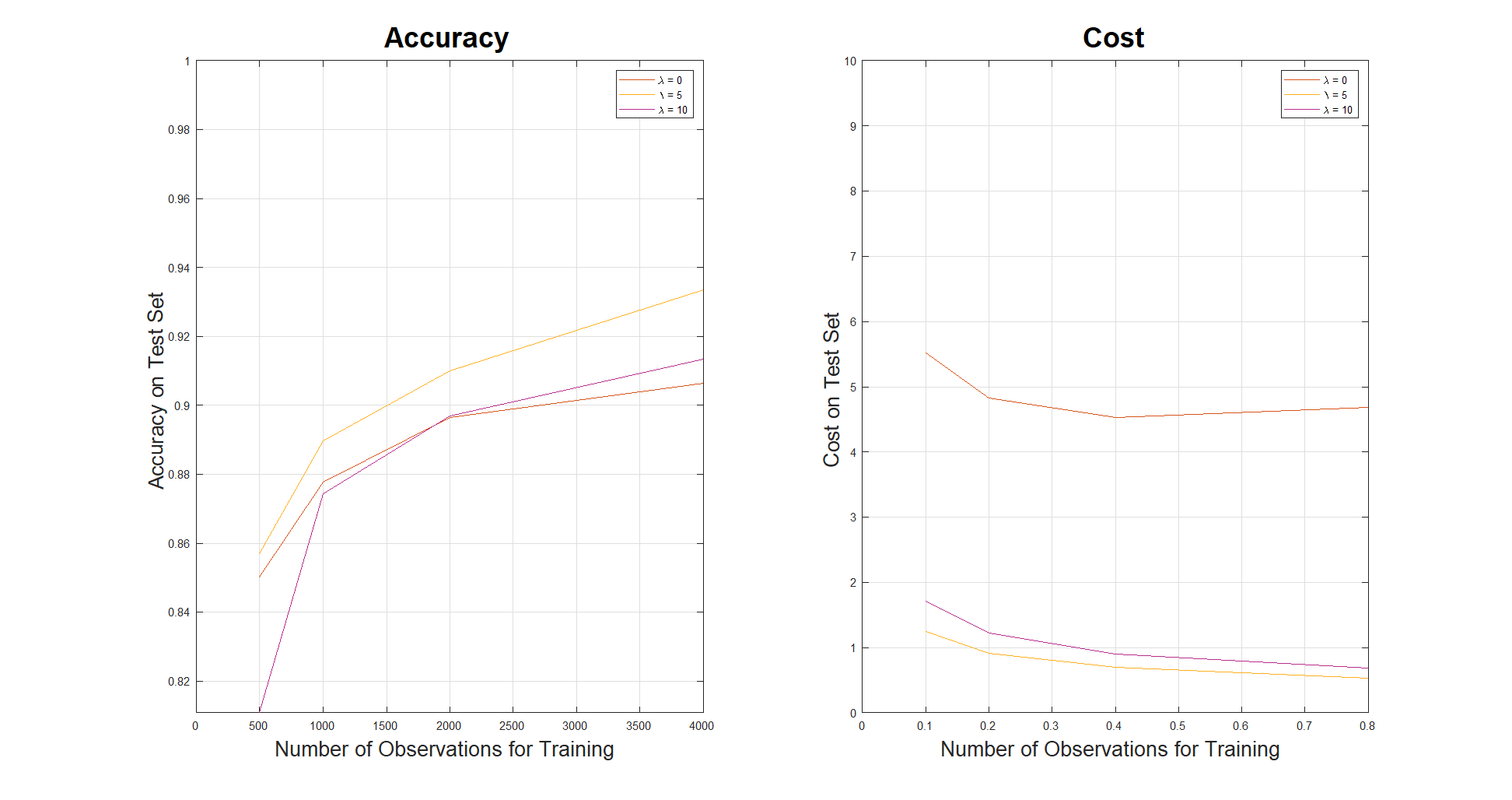}%
\end{center}
\end{figure}
First, we see that even 500 training samples produce sufficiently good
training in the sense that the teacher correctly classifies more than 80\% of
the test samples, while adding the other 3500 training samples increases the
accuracy by another 10--15\%.

Second, the performance of the teacher is visibly affected by the
regularization constant $\lambda$ that controls how a neural network fits to
the data. We consider three regularization levels: $\lambda=0,5,10$. A\ large
value of $\lambda$ penalizes large weight values; it results in a more general
model that possibly underfits the data (if $\lambda$ is big enough), leading
to a high bias. A\ small value of $\lambda$ allows large weight values; it
results in a more specific model that possibly overfits the data (if $\lambda$
is small enough), leading to a high variance. When $\lambda$ is 0, the
training algorithm does not penalize large weight numbers. An overfit model
will do badly on unseen samples, while an underfit model will do badly
overall. We see that $\lambda=5$\ delivers better results than $\lambda=0$
(underfit) and $\lambda=10$ (overfit) in terms of accuracy and cost.

\subsubsection{Training the student network}

The student's network differs from the teacher's network in two respects:
First, the student can learn either from the data or from the trained teacher,
and second, its network is trained by stochastic gradient descent, which
allows us to assess the learning speed. In all experiments, we set the
student's regularization $\lambda$ equal to 0 because the student is not
trying to generalize anything;\ rather, it is simply trying to learn.

\paragraph{Learning from a teacher versus learning from data.}

We consider two alternative learning procedures for the student: one is from
the data and the other is from the teacher. In the former case, the student is
an independent learner trained on the student training samples $X_{student}$
and the corresponding actual output $y_{student}$. In the latter case, the
student\ is trained on the same samples $X_{student}$ but uses the teacher's
predicted output $\hat{y}_{student}$.

What is the difference between $y_{student}$ and $\hat{y}_{student}$? In the
former case, we assert that $y_{student}$ represents a certain digit and does
not represent all of the other digits. As a result, each $y_{student}$ is a
$10\times1$ vector that has 9 zero values and 1 unit value. For example, the
$y_{student}$ vector is (1,0,0,0,0,0,0,0,0,0) if the digit is a 0. Another
possible interpretation is that the image is the given digit with probability
1 and any other digit with probability 0.

In turn, the teacher's predicted output $\hat{y}_{student}$ is a $10\times1$
vector where each element represents the probability that the image is a
certain digit as perceived by the trained teacher. Thus, $\hat{y}_{student}$
has same structure as $y_{student}$ but has all nonzero (positive) elements
$(p_{1},p_{2},...,p_{10})$ normalized to one by $\sum_{i=1}^{10}p_{i}=1$. We
will not round $\hat{y}_{student}$ to the most probable output because we
imagine that the teacher considers various alternatives rather than just one
answer and informs the student about the corresponding probability distribution.

\paragraph{Gradient descent and the learning rate.}

We will train a student neural network using stochastic gradient descent. The
student will train on one sample at a time: for each sample, we find the
gradient (partial derivatives of weights)\ based on the current sample. Then,
we update the neural network weights based on this gradient. The update is
performed according to the gradient descent rule:%
\begin{equation}
\theta \longleftarrow \theta-\alpha \times \frac{\partial J}{\partial \theta},
\label{B}%
\end{equation}
where $\alpha$ is a learning rate. This parameter describes how fast the
student learns from the current sample. I\ use two values of $\alpha$,
namely,\ 0.1 and 0.5. In my model, a high learning rate $\alpha=0.5$
represents a high-ability student, and a low learning rate $\alpha=0.1$
corresponds to a low-ability student. Models with a high learning rate may
converge quicker but are more prone to fluctuations and instability.

\section{The role of a teacher in the learning process}

In Section 2, we reviewed evidence from modern psychological literature about
the determinant of successful human learning. In Section 3, we designed a
computer simulation in the way that allows to capture some of these
determinants: For the teacher, we account for the "expert-novice" effect by
varying the number of handwritten samples that are used for training from 500
to 3,500; we model the teacher's ability to categorize information into big
ideas by varying the level of regularization $\lambda=0,5,10$; and we
construct examples of "bad" teachers with inadequate (biased) perception of
the learning process by training it on biased data sets, such as too hard or
too easy handwriting samples. Below, we assess how the size of a teacher's
training set, the level of regularization and the training set bias affect the
the learning outcomes.

\subsection{The importance of big ideas}

I first consider a student whose learning rate in the gradient descent rule
(\ref{B}) is equal to $\alpha=0.5$. I use 500 random handwritten samples for
training the student. (For the teacher, I\ consider three values of $\lambda$,
namely,\ 0, 5 and 10; and I use 500 handwritten samples for its training). The
results are shown in Figure 2.%

\begin{figure}

\begin{center}
\includegraphics[
height=3.334in,
width=6.0287in
]%
{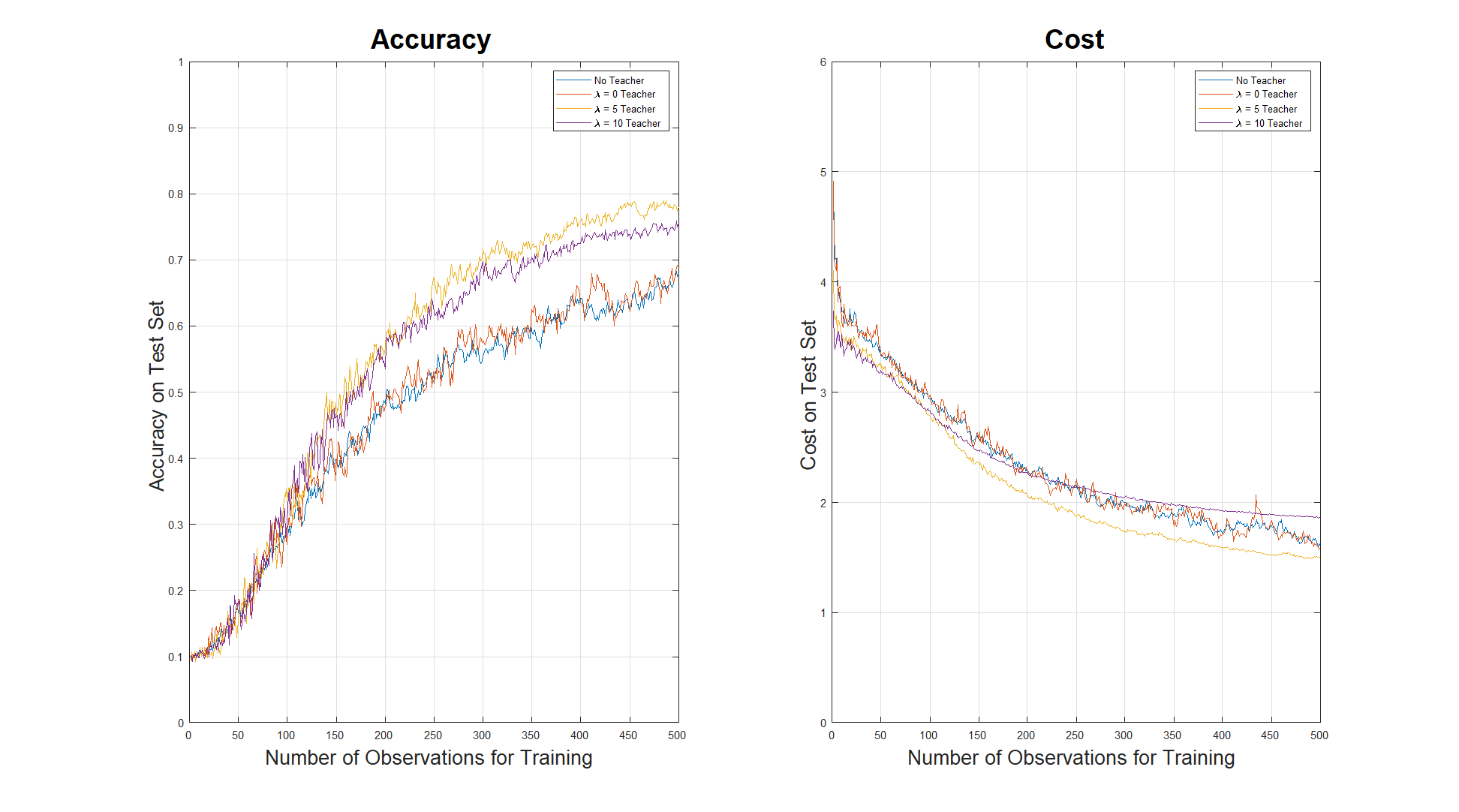}%
\end{center}
\end{figure}
\qquad

In the figure, we see that the highest predictive accuracy and lowest cost is
obtained when the student learns from the teacher with $\lambda=5$ and
$\lambda=10$. \textit{That is, the student learns faster from the
appropriately regularized teachers than both from the data and from overfit
teachers.}

Why is learning from an appropriately regularized teacher more effective than
learning from the data? This is because the teacher transmits to the student
its life-long experience rather than just one data point. The teacher has
analyzed many samples; it has distinguished the trends, and it now supplies to
the student the probabilities that a given image is every single possible
digit rather than just one correct digit value.

Why is the effectiveness of learning from an unregularized teacher lower and
comparable to learning from the data? When $\lambda=0$, the teacher does not
try to generalize the tendencies but tries to simply find parameters that
explain its own sample. Such an overfit teacher understands how to deal with
its own samples but gives a bad advice on other, unseen samples. Because there
is no overlap between the teacher's training set and the students' training
set, the effectiveness of learning from such a teacher is similar to the one
of the raw data.

As we argued in Section 2, modern psychological studies highlight the
importance of providing students with learning experiences that are designed
to enhance the student's abilities to recognize meaningful patterns of
information. The knowledge of expert teachers is not simply a list of facts
and formulas that are relevant to their domain; instead, their knowledge is
organized around core concepts or \textquotedblleft big
ideas\textquotedblright \ that guide their thinking about their domains. This
is exactly what the regularization parameter represents in our analysis.

\subsection{Novice teacher versus experienced teacher}

In the previous section, we analyzed learning from the teacher that was
trained with 500 samples. But what happens if we increase the teacher's
training set? Would the student perform better when learning from a more
trained teacher? In Appendix A.1, we show the results for a very experienced
teacher trained with 3,500 samples, and we obtain results that are very
similar to those in our baseline Figure 1. Why does giving the teacher more
samples not lead to visibly better learning outcomes? In our experiments, the
student is trained on 500 samples and reaches predictive accuracy of 70\% at
the end of the training. In that range, the student's knowledge is not yet
well-refined, so its learning progress does not considerably depend on whether
the teacher has the accuracy of 85\% or 95\%. For example, the learning of
students in an 8th grade algebra class may not critically depend on whether or
not their teacher took advanced Ph.D. classes in topology and multivariate calculus.

However, when the training of the teacher becomes insufficient relative to the
students level, the effectiveness of teaching begins to suffer. As an
illustration, in Figure 3, we consider an underqualified teacher that is
trained on just 100 samples.%
\begin{figure}[H]

\begin{center}
\includegraphics[
height=3.334in,
width=6.0287in
]%
{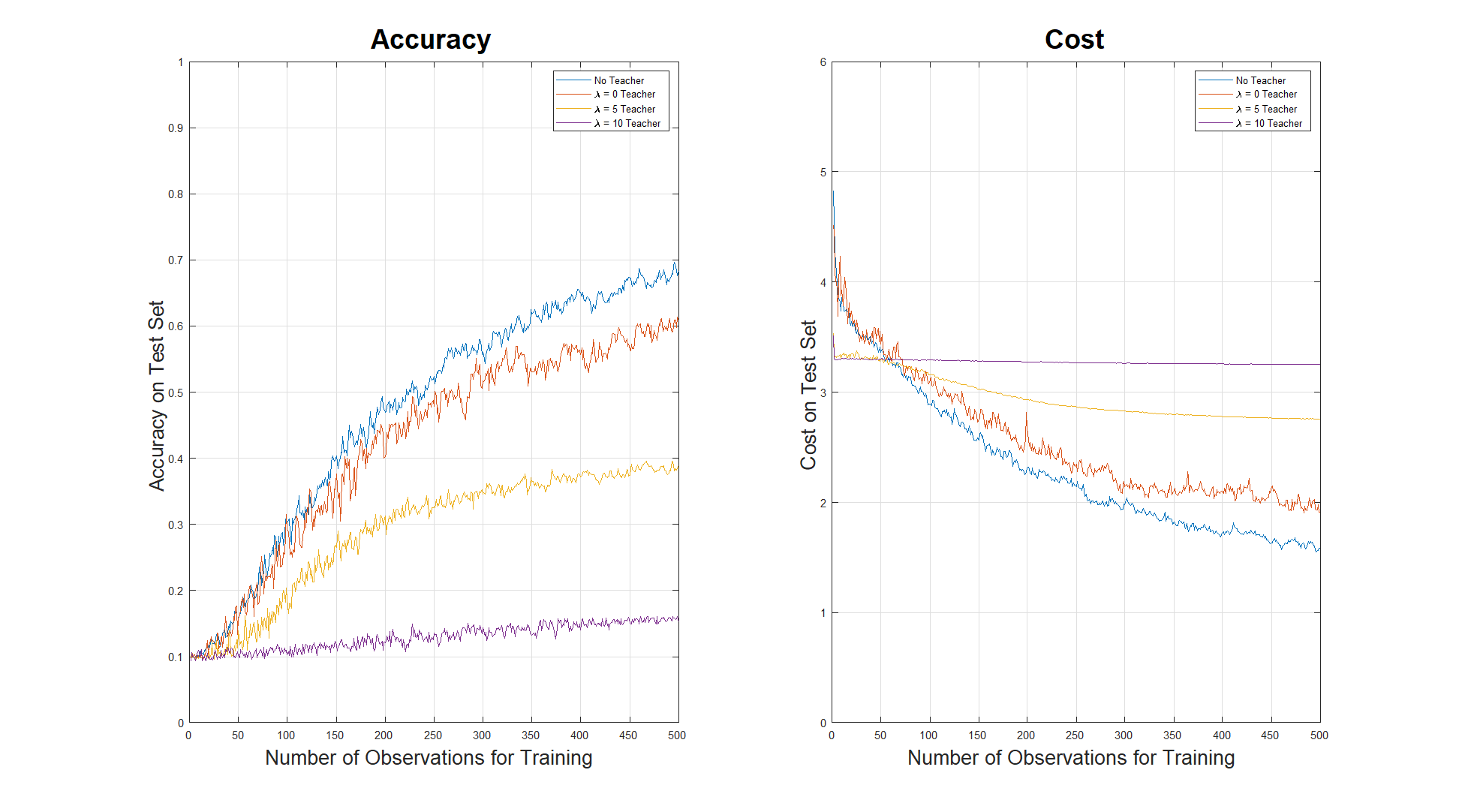}%
\end{center}
\end{figure}

In that case, learning from the data is faster than learning from the teacher,
which leads us to the following conclusion. \textit{The success of learning
depends on the qualification of the teacher relative to the qualification of
the student. Overqualification of the teacher does not improve learning
outcomes but underqualification worsens such outcomes.}

In fact, the student that is learning from the data will continue to progress.
In turn, the students that learn from the teacher are bounded by the
predictive accuracy that the teacher's training guarantees. Recall that a
higher level of regularization $\lambda$ implies that the teacher tries to
generalize the data and to identify a smooth trend. In that case, the teacher
has poor knowledge and insufficient experience to make generalizations -- just
100 samples -- and its poor judgement conditions the poor progress of the student.

\subsection{The teacher with a biased perception of learning process (bad
teacher)}

In some experiments, I observed that atypical handwritten samples disrupt the
training of the student. To assess the effect of such disruptions, I construct
two sets of atypical data: one set includes handwriting samples that are
hardest to classify and the other set contains handwriting samples that are
the easiest to classify. For this experiment, I train a neural network on the
set of 3,500 observations until convergence and pick out the two tails data
sets: 500 observations with the highest cost and 500 observations with the
lowest costs.

In Figure 4, I illustrate how well the students learn if their teacher has a
bias because it is trained on the set with hard handwriting.%

\begin{figure}[H]

\begin{center}
\includegraphics[
height=3.334in,
width=6.0287in
]%
{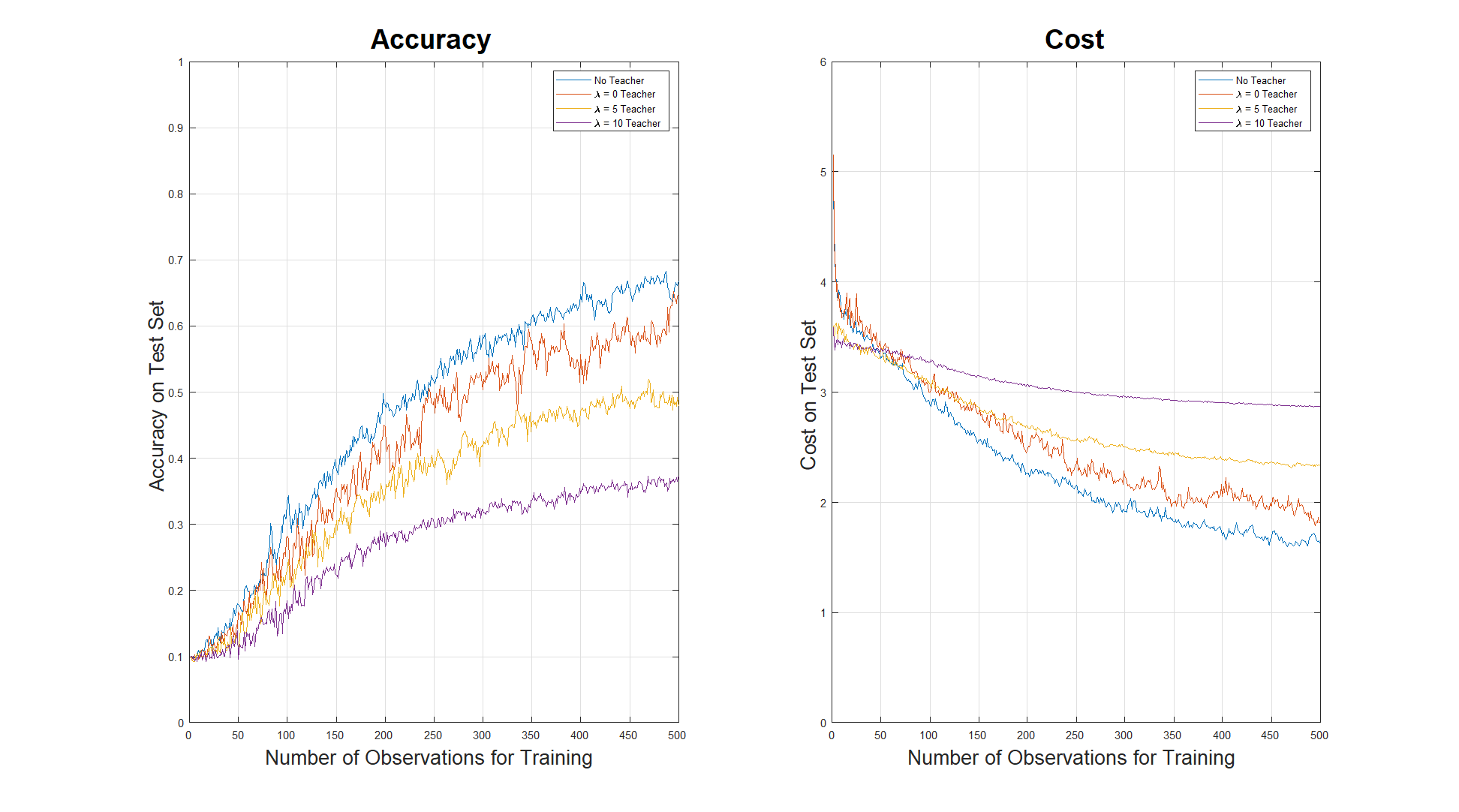}%
\end{center}
\end{figure}

We observe that the student that learns from the data performs better than the
student taught by a teacher trained on hard data. In Appendix A.2, I show that
the results are similar when the teacher is trained on 500 easiest samples,
which leads us to the following conclusion: \textit{Learning from a teacher
with a biased perception of the learning process is less effective than
learning from the data directly, and regularization only increases the bias.
}The teacher that is trained on atypical data has an inadequate perception
about typical handwriting. It becomes even worse when such a teacher tries to
distinguish trends via a higher level of regularization. The distinguished
trend reflects the hard data and not the typical data, so the teacher gives
systematically bad advice to the students. In contrast, the student that
learns from the data observes the typical samples and identifies the trend correctly.

\section{The learning success depends on the students too}

In the previous section, I show that the training and analytical abilities of
the teacher (i.e.. its capacity to generalize the trends) are critical for the
learning success of the students. But the evidence outlined in Section 2 show
that the student's characteristics play an important role in the learning
outcomes as well, in particular, their abilities and socioeconomic background.
For the student, we control for the level of abilities by varying the learning
rate $\alpha$; and I model the student's biased experience (due to, e.g.,
poverty and trauma) by using biased training data sets that do not coincide
with the experience of a typical student. Below, I assess the role of the
student's characteristics for the learning outcomes.

\subsection{A low ability student}

In Section 2.1, we considered a student with learning rate in the gradient
descent rule (\ref{B}) equal to $\alpha=0.5$. We use 500 random (typical)
handwritten samples for training the student; see Figure 2. We now consider
the student whose learning abilities are lower $\alpha=0.1$; see Figure 5.%

\begin{figure}[H]

\begin{center}
\includegraphics[
height=3.334in,
width=6.0287in
]%
{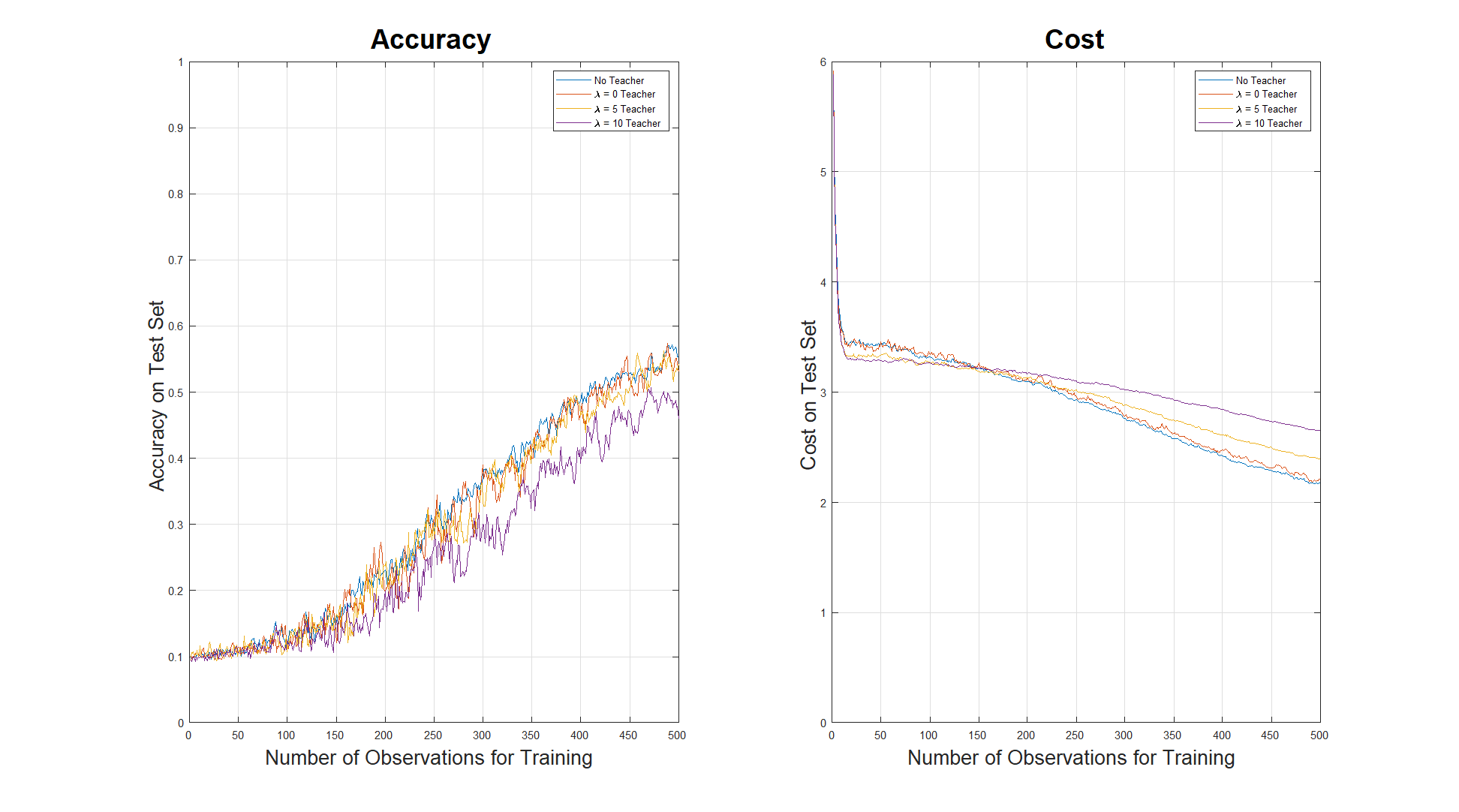}%
\end{center}
\end{figure}

Here, all training schemes lead to similar learning outcomes, except of that
with an overfit teacher $\lambda=10$ that produces the worst results in terms
of accuracy and costs. \textit{Thus, in our experiment, low-ability students
benefit less from a teacher than high-ability students.} This finding accords
remarkably with the evidence from Kulik and Kulik (1982)\textbf{. }

\subsection{A student with a bias}

Let us finally consider an experiment in which the teacher is trained on the
typical data and students are trained on atypical (hard, easy) data. In Figure
6, we show the case when the students are trained with the hard samples.
\begin{figure}[H]

\begin{center}
\includegraphics[
height=3.334in,
width=6.0287in
]%
{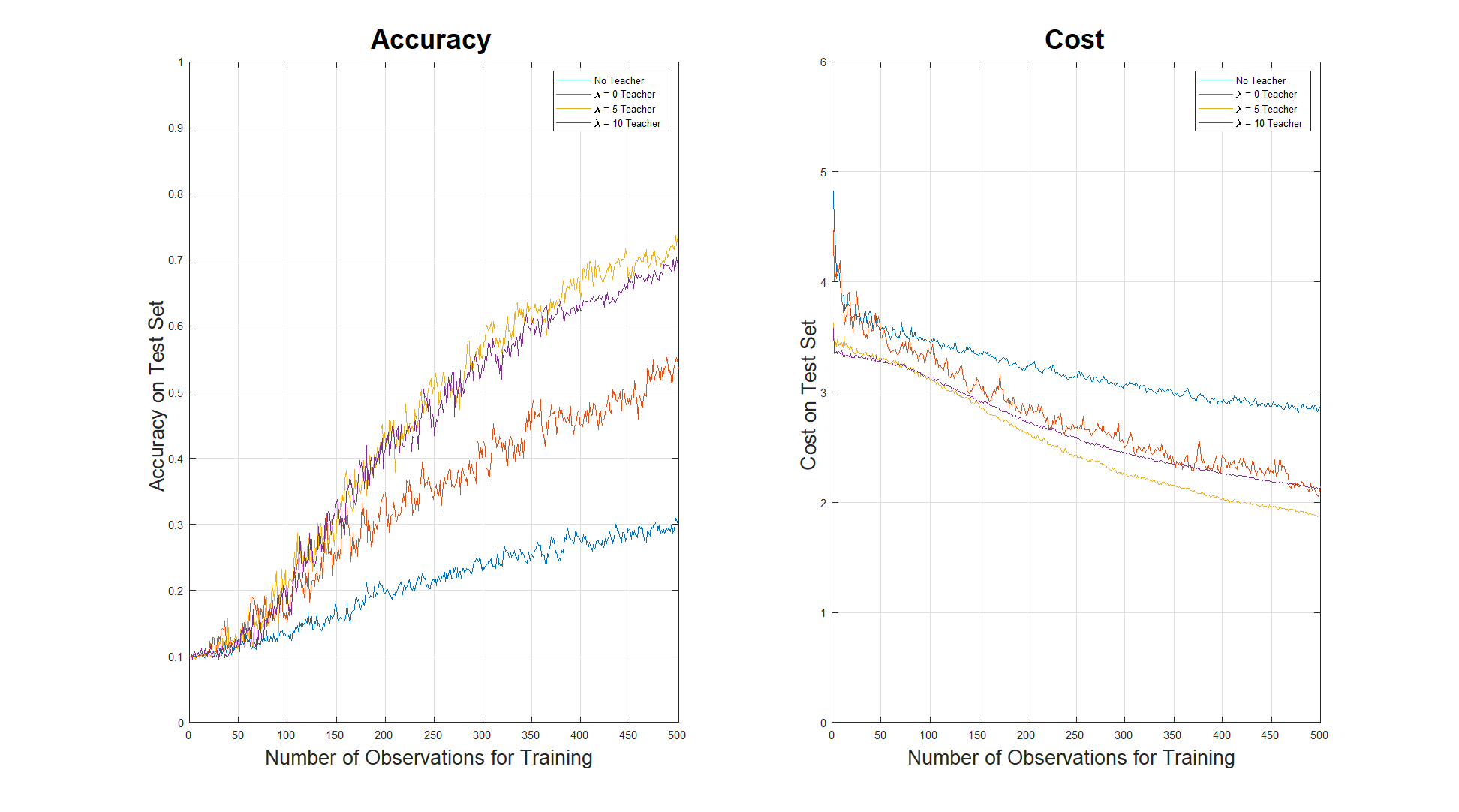}%
\end{center}
\end{figure}

We now observe that the students that learn from hard samples benefit greatly
from the teacher, in particular, from the appropriately regularized teacher.
In Appendix A.3, I show that we observe the same regularities when the
students are learning from 500 easiest samples, which leads us to conclude the
following: \textit{Learning from a teacher with typical experience is better
than learning from biased data; moderate regularization can increase the
effectiveness of the teacher. }

There is simple intuition behind this result. The student that learns from
biased samples has inadequate experience and makes mistakes when tested on the
typical samples. But the teacher that was trained on typical samples gives
correct advice to the student even when the samples are atypical, which
mitigates the effect of biased samples on the learning process. Regularization
reinforces the learning process as in our benchmark case. Thus, our computer
simulation grossly accords with the idea that the educator can become the
missing person for those students whose own experience is contaminated by
negative experience (e.g., poverty and trauma) by overriding those inadequate
responses that the students observed from their own biased socioeconomic
experience (as shown by the finding in Section 2).

\section{Conclusion}

The main message of the present paper is that we can gain understanding into
the determinants of human learning by simulating the interactions of
artificial intelligence. The advantage of computer simulation compared to
experimentation with human subjects is that we have a better control over the
learning experiments, we understand how the learning process depends on the
model's parameters, and we can re-run the experiments under different
parameterizations as many times as needed. This approach made it possible to
illustrate the importance of big ideas in the learning process and to
successfully model a variety of interesting learning situations such as
expert-novice teachers, high-low ability students and atypical learning
experience, among others. My results are suggestive and provide explanation to
the real world learning experiences documented in the modern psychological literature.

\section*{Appendix}

This appendix contains the results of supplementary experiments.

\subsection{Appendix A1}

In Section 3.1, we analyzed the outcome of learning when the teacher was
trained with 500 samples. In Figure 7, we provide the sensitivity results in
which the teacher is trained with a larger number of observations, namely,
3,500. We observe that the outcomes of learning process in Figures 2 and 7 are
very similar, which indicates that excessive training of the teacher does not
improve the effectiveness of learning.%

\begin{figure}[H]

\begin{center}
\includegraphics[
height=3.334in,
width=6.0287in
]%
{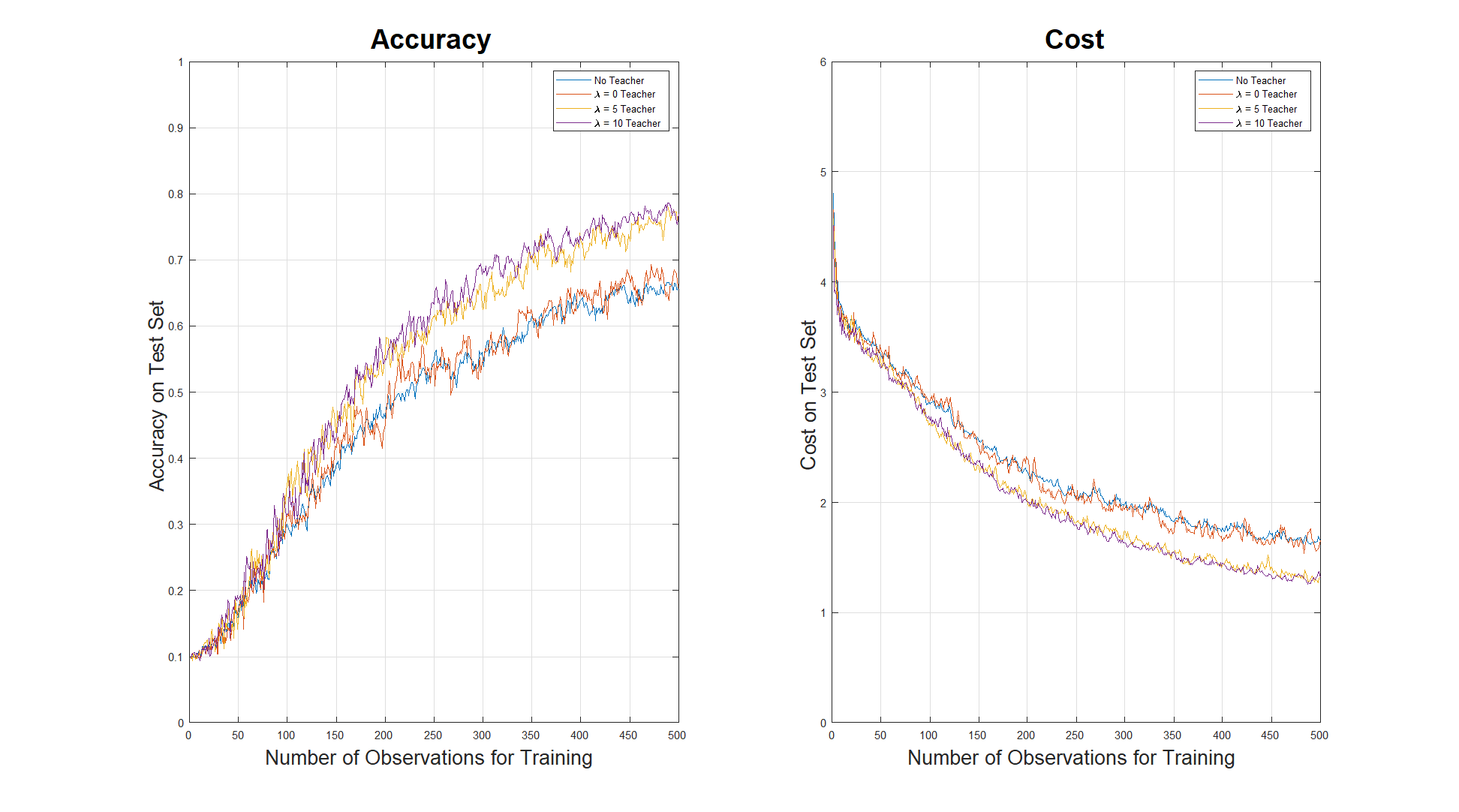}%
\end{center}
\end{figure}

\subsection{Appendix A2}

In Section 3.3, we show that training the teacher on the hard samples reduces
the effectiveness of the teacher; see Figure 4. We now consider the opposite
case when the teacher is trained on the easiest samples; see Figure 8. We
observe that learning from the data is more effective than learning from such
a teacher. This leads us to the conclusion in the main text: a bias (in either
direction) reduces the effectiveness of teaching %

\begin{figure}[H]

\begin{center}
\includegraphics[
height=3.334in,
width=6.0287in
]%
{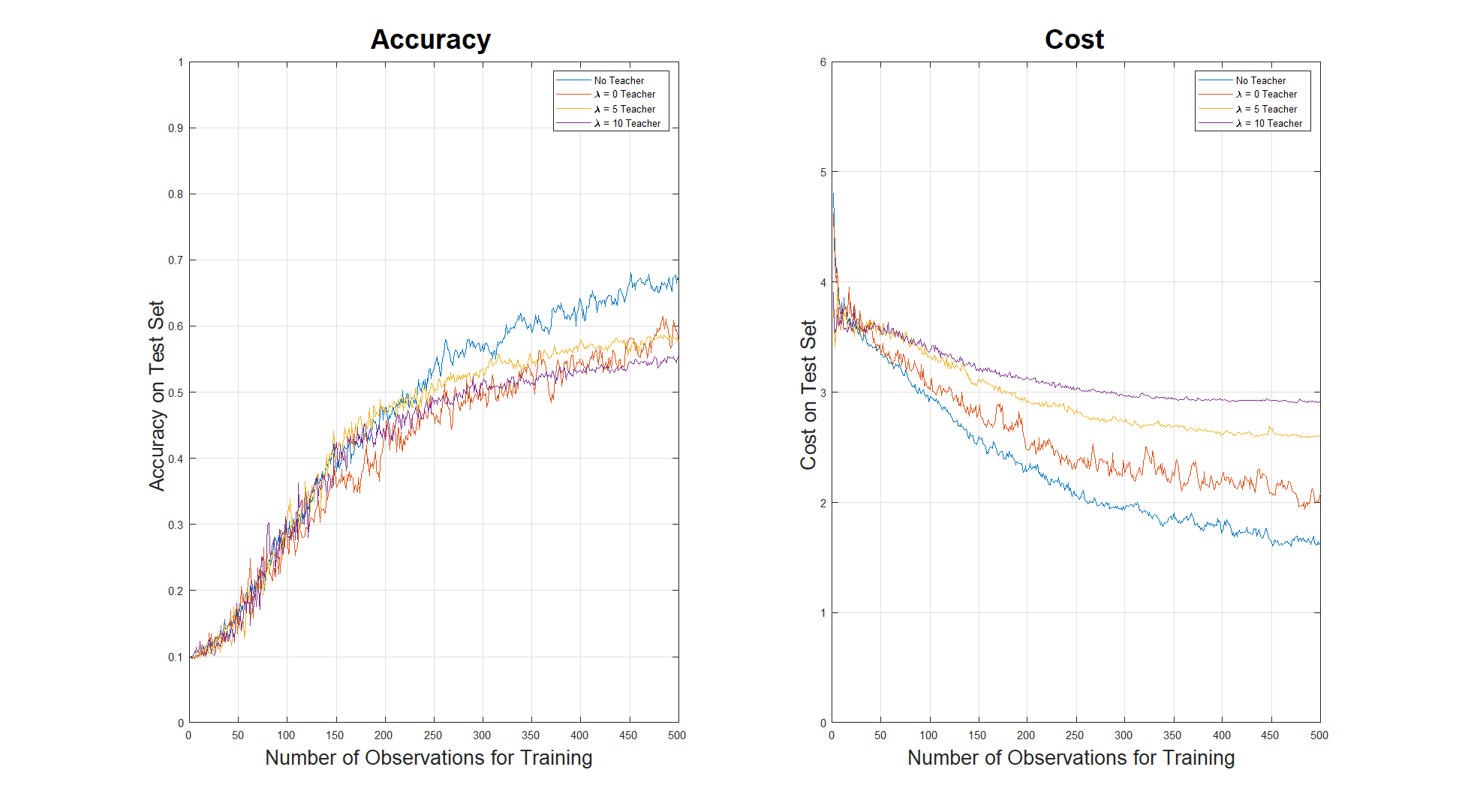}%
\end{center}
\end{figure}

\subsection{Appendix A3}

In Section 4.2, we consider the student that is trained on excessively hard
samples. We now consider the remaining case of the student that is trained
with excessively easy samples; see Figure 9.
\begin{figure}[H]

\begin{center}
\includegraphics[
height=3.334in,
width=6.0287in
]%
{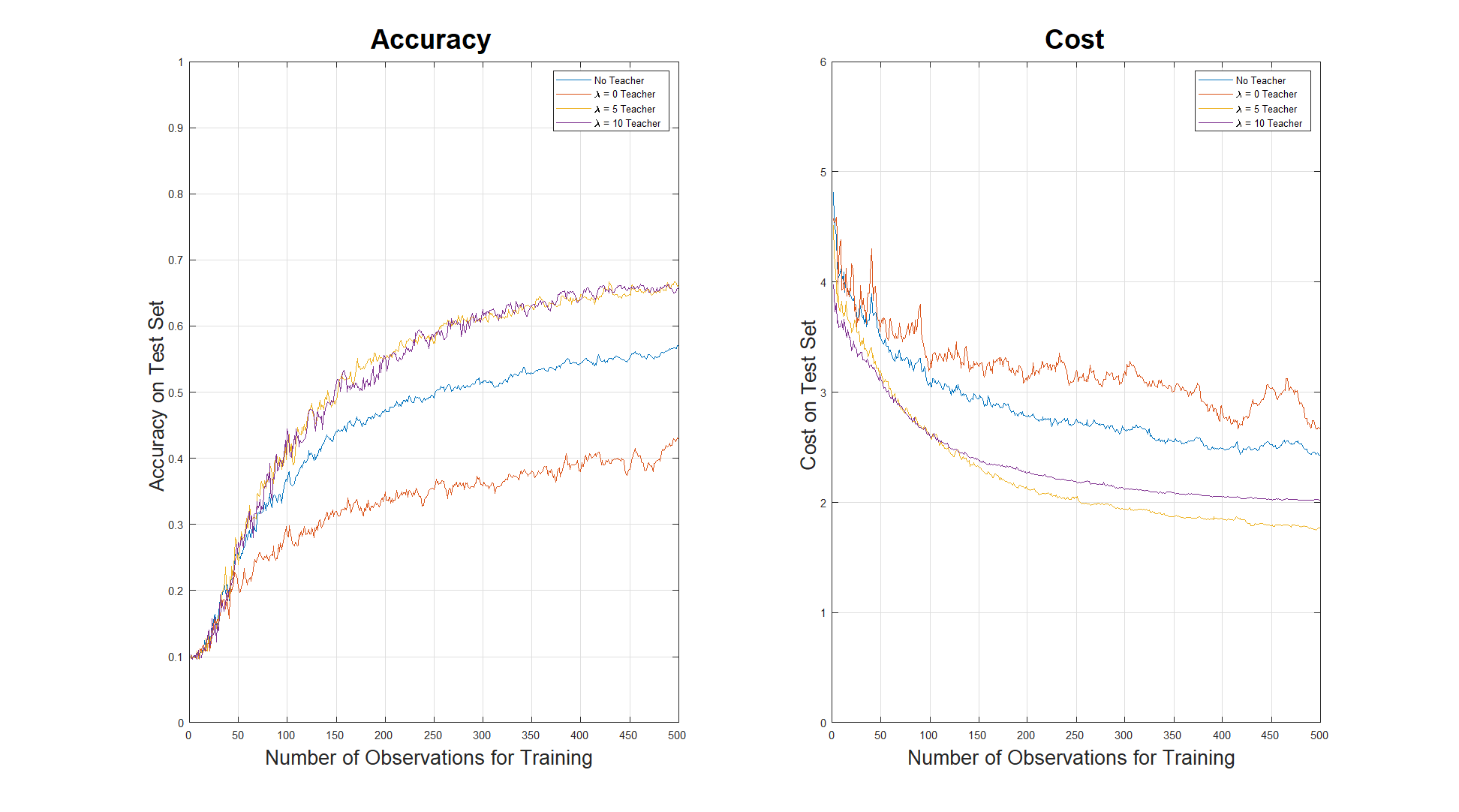}%
\end{center}
\end{figure}

Comparing Figures 5 and 9, we observe that in both cases (hard and easy
samples), learning from the teacher dominates learning from the data. The
difference is that unregularized (overfit) teacher performs poorly. This is
because such a teacher learned to reproduce its specific typical sample rather
than learning the trends. The advice of such a teacher is generally poor, and
it is especially poor for the biased sample that the student faces.

\end{document}